\documentclass[preprint, 10pt]{elsarticle}
\usepackage[utf8]{inputenc}

\usepackage[left=2.5cm,right=2.5cm,top=3cm,bottom=3cm]{geometry} 
\usepackage{setspace} 
\usepackage{lineno}  
\usepackage{lipsum}   
\usepackage{enumitem}

\usepackage{amsmath}    
\usepackage{amsfonts}  
\usepackage{bm}         
\usepackage[mathscr]{euscript} 
\DeclareMathAlphabet{\mathpzc}{OT1}{pzc}{m}{it}
 
\usepackage{tikz}
\usetikzlibrary{arrows.meta, positioning}

\usepackage{graphicx}      
\usepackage{threeparttable} 
\usepackage{subcaption}    
\usepackage{booktabs}      
\usepackage{tabularx}      
\usepackage{multirow}      
\usepackage{makecell}      
\usepackage[table]{xcolor} 
\usepackage{svg}           
\usepackage{epstopdf}      
\usepackage{picinpar}      

\usepackage{comment}
\usepackage{array,booktabs,tabularx,threeparttable,xcolor}

\usepackage[section]{placeins} 
\biboptions{authoryear,compress}

\newdefinition{assumption}{Assumption}
\newdefinition{definition}{Definition}
\newdefinition{remark}{Remark}
\newproof{proof}{Proof}

\let\OLDthebibliography\thebibliography
\renewcommand\thebibliography[1]{
  \OLDthebibliography{#1}
  \setlength{\parskip}{0pt}
  \setlength{\itemsep}{0pt plus 0.3ex}
}

\usepackage[colorlinks=true, linkcolor=blue, citecolor=blue, urlcolor=blue]{hyperref}

\usepackage{cleveref}  

\makeatletter
\def\ps@pprintTitle{
 \let\@oddhead\@empty
 \let\@evenhead\@empty
 \let\@oddfoot\@empty
 \let\@evenfoot\@oddfoot}
\makeatother

\begin{document}
\begin{frontmatter}

\title{Learning from Risk: LLM-Guided Generation of Safety-Critical Scenarios with Prior Knowledge}

\author[a]{Yuhang Wang}
\ead{wangyuhang22@mails.ucas.ac.cn}

\author[b,c]{Heye Huang\corref{cor1}}
\ead{heyeh@mit.edu}

\author[d]{Zhenhua Xu}
\ead{zhenhuaxu@tsinghua.edu.cn}

\author[b,c]{Kailai Sun}
\ead{skl24@mit.edu}

\author[b,c]{Baoshen Guo}
\ead{baoshen@mit.edu}

\author[b,c]{Jinhua Zhao}
\ead{jinhua@mit.edu}

\address[a]{Chinese Academy of Sciences, China}
\address[b]{Singapore-MIT Alliance for Research and Technology Centre (SMART), Singapore}
\address[c]{Department of Urban Studies and Planning, Massachusetts Institute of Technology, USA}
\address[d]{School of Vehicle and Mobility, Tsinghua University, China}
\cortext[cor1]{Corresponding author}

\begin{abstract}
%

Autonomous driving faces critical challenges in rare long-tail events and complex multi-agent interactions, which are scarce in real-world data yet essential for robust safety validation. This paper presents a high-fidelity scenario generation framework that integrates a conditional variational autoencoder (CVAE) with a large language model (LLM). The CVAE encodes historical trajectories and map information from large-scale naturalistic datasets to learn latent traffic structures, enabling the generation of physically consistent base scenarios. Building on this, the LLM acts as an adversarial reasoning engine, parsing unstructured scene descriptions into domain-specific loss functions and dynamically guiding scenario generation across varying risk levels. This knowledge-driven optimization balances realism with controllability, ensuring that generated scenarios remain both plausible and risk-sensitive. Extensive experiments in CARLA and SMARTS demonstrate that our framework substantially increases the coverage of high-risk and long-tail events, improves consistency between simulated and real-world traffic distributions, and exposes autonomous driving systems to interactions that are significantly more challenging than those produced by existing rule- or data-driven methods. These results establish a new pathway for safety validation, enabling principled stress-testing of autonomous systems under rare but consequential events. Code is available at \url{https://github.com/echoleaeperw/LRF}

\end{abstract}


\end{frontmatter}

%
\section{Introduction}
\label{intro}
%

The safety and reliability of autonomous driving depend on rigorous validation under diverse test conditions, especially in high-risk, highly interactive, and safety-critical scenarios~\citep{wang2021towards,hossain2025taxonomy}. Yet such events are extremely scarce in real-world datasets, creating a persistent gap between development testing and deployment needs. Simulation-based methods provide an effective alternative by generating large numbers of rare and adversarial environments, thereby alleviating data scarcity and enabling controlled safety evaluation~\citep{huang2020integrated}. However, existing scenario generation approaches face key limitations: (1) rule-based or parameterized methods offer interpretability and control but lack behavioral diversity, failing to reproduce the complexity of real-world human driving~\citep{zhao2016accelerated, sun2021scenario,yang2023adaptive}; (2) data-driven replay or imitation methods can reproduce historical interactions but cannot synthesize unseen long-tail risks~\citep{suo2021trafficsim, bergamini2021simnet}; and (3) recent generative models, though more flexible, still struggle to jointly ensure controllability, realism, and scalability~\citep{stoler2025seal, shao2024lmdrive,zhang2024chatscene}.

To address these challenges, this paper proposes a risk knowledge–guided traffic scene generation framework that integrates a Conditional Variational Autoencoder (CVAE) with a Large Language Model (LLM). Unlike prior works that merely sample or replay specific risky cases, the proposed framework establishes a general and controllable pipeline for synthesizing diverse safety-critical scenarios under varying risk conditions. The CVAE learns latent spatiotemporal representations from real-world trajectories and maps to generate physically coherent base scenes, while the LLM acts as a knowledge-driven controller that interprets scene semantics, analyzes multi-agent risk interactions, and dynamically adjusts optimization objectives to guide the generation toward desired levels of behavioral complexity and risk exposure. Through this dual mechanism and a knowledge-guided loss adaptation process, the framework enables interpretable and controllable generation across low-, high-, and long-tail risk regimes, effectively bridging the gap between simulation and real-world safety validation. The main contributions are as follows:

\begin{enumerate}

\item We propose a high-fidelity safety-critical scenario generation framework that integrates CVAE-based motion learning with LLM-guided optimization. This design enables a general and controllable pipeline for synthesizing diverse, risk-sensitive driving scenarios.

\item We introduce a knowledge-driven loss adaptation mechanism, where the LLM interprets risk semantics and dynamically adjusts optimization objectives. This ensures physical plausibility, behavioral consistency, and interpretable risk levels throughout the generation process.

\item We present a cross-risk distribution strategy that systematically covers low-, high-, and long-tail scenarios. It enhances long-tail representation and effectively bridges the sim-to-real gap in autonomous driving validation.

\end{enumerate}

%
\section{Related Works}
\label{rw}
%

Existing research on traffic scenario generation can be broadly categorized into three classes ~\citep{chen2024data,gao2025foundation}: (1) rule-based or parameterized sampling methods, (2) data-driven simulation methods, and (3) generative and adversarial modeling methods. Rule-based and sampling approaches emphasize controllability and interpretability but lack diversity. Data-driven approaches improve realism by leveraging large-scale datasets but struggle to generate unseen critical events. More recently, generative and adversarial approaches, including diffusion-based and LLM-driven frameworks, have advanced controllability and risk coverage, though they remain challenged by efficiency and scalability.

\textbf{(1) Rule-based and parameterized sampling methods.}
Early approaches to traffic scenario generation relied on rule-based log replay or parameterized sampling. These methods offered strong interpretability and controllability, since human designers could enforce traffic rules or specify event triggers such as cut-ins or sudden stops~\citep{liu2024controllable, deng2025target}. Classical simulators such as SUMO and CARLA provided configurable environments where adversarial agents could be inserted or trajectories perturbed to stress-test a system. Importance sampling techniques and surrogate models were also developed to accelerate the evaluation of rare safety-critical events by resampling from manually defined parameter spaces~\citep{zhao2016accelerated, sun2021scenario}. 
While these approaches achieved efficiency and reproducibility, their reliance on pre-specified rules and parameters constrained their ability to represent the diversity and stochasticity of real-world human driving. 
The behavioral space they explore is therefore bounded by designer assumptions, and no genuinely novel interactions can emerge once the rules or parameter ranges are fixed. In addition, most rule-based frameworks fail to preserve the joint dynamics among multiple agents, leading to unrealistic or oversimplified responses when scaled to dense traffic. These handcrafted designs also struggle to reflect the subtle variability in driver intent and reaction timing that strongly influences real-world risk. As a result, such approaches often overfit to synthetic corner cases rather than exposing realistic near-miss situations, making it difficult to evaluate how autonomous policies perform under nuanced uncertainty. Consequently, rule-based and sampling methods frequently miss the long-tail interactions that are most critical for safety validation~\citep{rempe2022generating, zhou2025crash}.

\textbf{(2) Data-driven simulation methods.}
With the availability of large-scale naturalistic driving datasets (e.g., nuScenes~\citep{caesar2020nuscenes}, Waymo Open Motion Dataset~\citep{sun2020scalability}, highD~\citep{krajewski2018highd}, inD~\citep{bock2020ind}), learning-based traffic simulation has emerged as a more realistic alternative. 
Methods such as TrafficSim~\citep{suo2021trafficsim} and SimNet~\citep{hennigh2021nvidia} learn multi-agent motion priors from logged trajectories, enabling the replay and synthesis of plausible interactions~\citep{chen2025rift, lin2025causal, wang2025traffic, SUN2025111265}. More recent efforts leverage generative adversarial networks and imitation learning to model complex traffic behaviors directly from data, allowing more dynamic and realistic multi-agent interactions~\citep{bergamini2021simnet}. 
These data-driven approaches significantly improve realism and diversity compared to rule-based methods. However, they often lack controllability, as generated behaviors are constrained by the distributions observed in training datasets~\citep{stoler2025seal}. 
This data dependency limits their ability to reproduce safety-critical edge cases that rarely occur in naturalistic data. In addition, models trained on specific regions or traffic conditions may generalize poorly to unseen environments, leading to bias and instability in simulation outcomes. The generated scenarios also tend to reproduce statistically frequent patterns rather than strategically valuable rare events. Furthermore, attempts to improve diversity through random perturbation or adversarial noise often degrade physical realism or disrupt coordination between agents. Consequently, purely data-driven simulations are insufficient for systematic testing under long-tail or adversarial conditions~\citep{liu2025rolling, aiersilan2025generating}.

\textbf{(3) Generative and adversarial modeling methods.}
To overcome the limitations of purely rule-based and data-driven approaches, generative models have been increasingly adopted for traffic scenario generation~\citep{sheng2025talk2traffic}. These approaches aim to balance realism, diversity, and controllability by learning from data while enabling the synthesis of unseen safety-critical situations.  

Latent-variable generative models such as STRIVE leverage graph-based conditional VAEs trained on real-world data to perturb existing trajectories in the latent space, creating accident-prone yet plausible scenarios~\citep{rempe2022generating}. By optimizing adversarial objectives, these models generate diverse collisions and identify planner weaknesses. Similar methods have employed GANs or probabilistic motion priors to enrich scenario diversity~\citep{feng2023dense}. However, their reliance on gradient-based optimization restricts scalability and often limits interpretability when extended to multi-agent interactions. In addition, the controllability of these models is typically constrained to a few predefined numerical indicators, making it difficult to specify semantic goals such as “aggressive cut-in” or “near-miss merging.”  

Diffusion-based models further enhance realism and controllability by gradually refining trajectories from noise to physically coherent motion~\citep{xu2025diffscene}. They provide strong generative capacity and can model complex, multi-modal distributions with high fidelity, producing continuous transitions between safe and unsafe behaviors. More recent studies have extended diffusion models with auxiliary guidance signals that encode scene semantics and geometric priors, enabling fine-grained control over trajectory realism and interaction intensity. These methods significantly expand the coverage of safety-critical conditions while maintaining naturalistic motion patterns. Nevertheless, diffusion-based models remain computationally expensive and require numerous denoising steps for each sample, which limits their scalability for large-scale testing and real-time simulation.  

Beyond diffusion and adversarial sampling, recent work explores using LLMs \citep{sun2025review} to guide the formation of loss or reward functions in reinforcement learning and traffic simulation. Liu et al. introduced an LLM-guided hierarchical reasoning framework with chain-of-thought prompting and Frenet-frame cost functions, improving the model’s understanding of spatial relationships and interaction intent~\citep{liu2024controllable}. These designs enhance semantic controllability, allowing structured manipulation of scenario difficulty, vehicle roles, and interaction types. Han et al. further demonstrated that LLMs can automatically generate and refine reward functions for highway driving through iterative reflection, achieving higher training success rates and reducing manual design effort~\citep{han2024generating}. Such approaches highlight the potential of language-driven supervision to shape optimization objectives in a human-interpretable and adaptive manner. However, when applied in isolation, LLM-based methods often lack grounding in physically consistent motion priors, leading to unstable or unrealistic behaviors under dense multi-agent conditions.

%
\section{Methodology}
\label{methodology}
%

\subsection{Overall Framework}

\begin{figure}[htbp]
    \centering
    \includegraphics[width=1.0\textwidth]{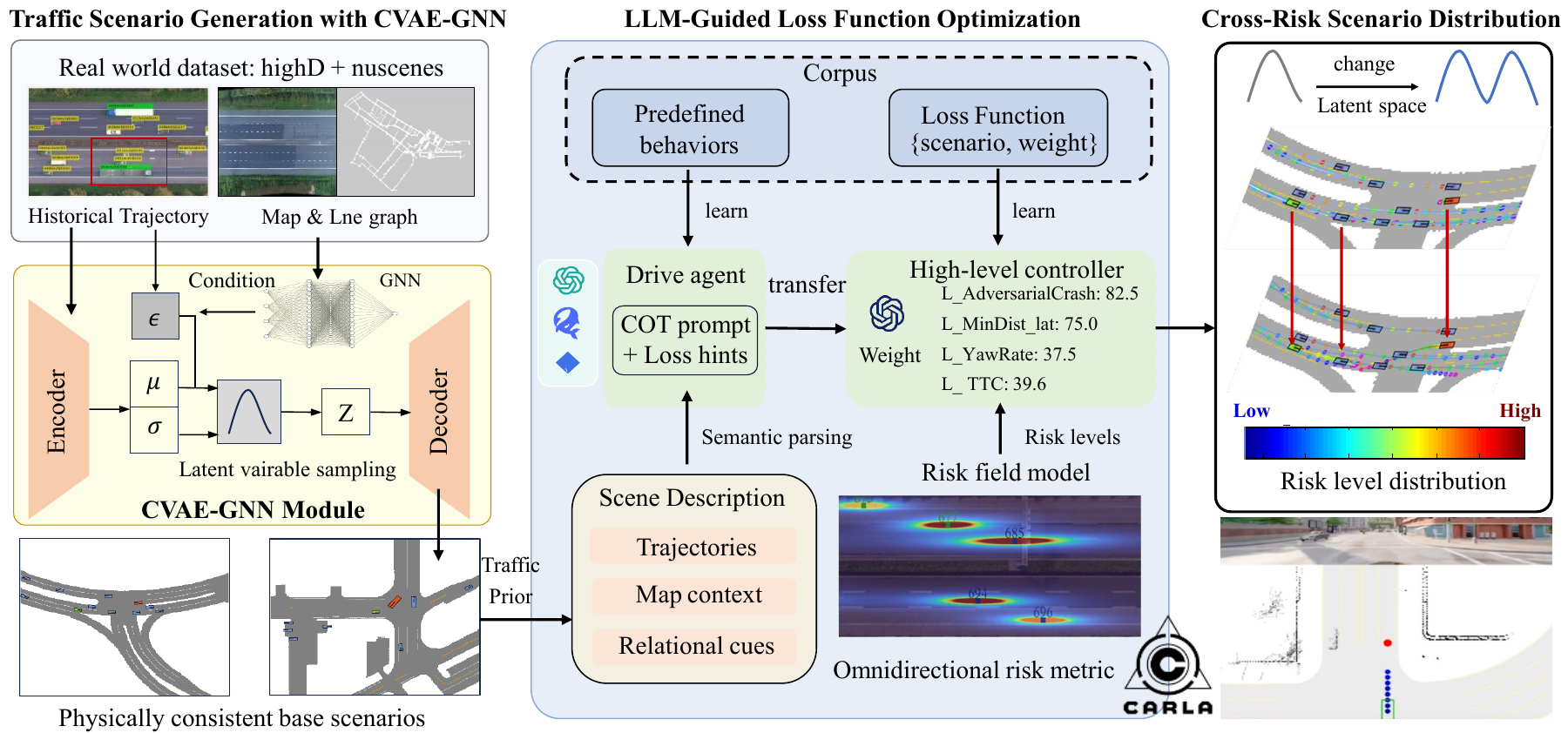}
    \caption{Overview of the proposed LLM-guided CVAE-GNN framework for safety-critical traffic scenario generation. 
    The left module learns motion priors from large-scale datasets (\textit{highD}, \textit{nuScenes}) and encodes multi-agent interactions into a latent variable $z$.
    The middle module parses scene semantics and employs chain-of-thought prompting to generate adaptive loss hints. 
    A high-level controller integrates these hints with a risk field model that computes omnidirectional risk metrics and gradient feedback, dynamically adjusting loss weights ($L_\text{AdversarialCrash}$, $L_\text{MinDist\_lat}$, $L_\text{YawRate}$, $L_\text{TTC}$).
    The right module reshapes the latent space to generate scenarios covering low-, medium-, and high-risk regimes, resulting in a continuous distribution of risk-aware driving interactions.}
    \label{fig:overall_framework}
\end{figure}

We propose a unified framework for generating safety-critical traffic scenarios that integrates data-driven motion priors with knowledge-guided reasoning. As illustrated in Figure~\ref{fig:overall_framework}, the framework consists of three interconnected modules.
(1) Traffic scenario generation with CVAE-GNN. Historical trajectories and high-definition map data from large-scale datasets such as highD and nuScenes are encoded by a graph-based conditional variational autoencoder. The encoder captures multi-agent interactions and lane topology to form a latent variable $z \sim \mathcal{N}(\mu, \sigma^2)$ representing stochastic traffic behavior, while the decoder reconstructs physically consistent base scenarios $\hat{Y}$ as realistic and diverse motion priors.
(2) LLM-guided loss function optimization. Structured scene descriptions containing trajectories, map context, and inter-agent relations are parsed by an LLM that identifies event types and potential risks through chain-of-thought reasoning. The LLM outputs adaptive loss weights (e.g., $L_\text{AdversarialCrash}$, $L_\text{MinDist\_lat}$, $L_\text{YawRate}$, $L_\text{TTC}$), which are iteratively refined via a physics-based risk field model to ensure semantic interpretability and dynamic stability.
(3) Cross-risk scenario distribution. The optimized latent space is reshaped to continuously span low-, medium-, and high-risk regimes. By modulating the learned distribution, the framework generates balanced scenarios across varying risk levels, enabling systematic stress-testing of autonomous driving policies under diverse, safety-critical conditions.
Overall, this framework transforms traditional data-driven replay into a knowledge-guided generative paradigm, unifying semantic reasoning with latent-space optimization for controllable, physically consistent, and risk-aware traffic scenario generation.

\subsection{Learning Traffic Priors with CVAE-GNN}

To generate realistic and diverse traffic scenarios, we construct a deep generative model that learns latent distributions from real traffic data and samples from them to synthesize new trajectories. Building on STRIVE~\citep{rempe2022generating}, this generative prior serves as the foundation for constructing safety-critical scenarios. By performing optimization in the latent space, the framework ensures that generated trajectories remain physically feasible while exhibiting controllable levels of interaction risk.

We adopt a CVAE as the core generator and integrate a graph neural network (GNN) into its encoder~\citep{rahmani2023graph} to capture complex multi-agent dependencies. In this formulation, vehicles are represented as nodes and their pairwise relations, such as distance, relative velocity, and potential conflict zones, are encoded as edges. Through message passing, the encoder aggregates relational information and produces a latent variable $z \sim \mathcal{N}(\mu, \sigma^2)$ that captures stochastic yet structured traffic dynamics. The decoder then reconstructs multi-agent trajectories that are spatially coherent and kinematically consistent. Modeling scenario generation as an optimization process in the latent space allows the framework to balance realism, behavioral diversity, and adversarial challenge, providing a principled way to represent safety-critical interactions across a continuous risk spectrum.
High-risk behaviors in naturalistic traffic—such as lane changes, merges, and aggressive maneuvers, occur in both urban and highway environments. To capture this diversity, our implementation is trained jointly on nuScenes and highD datasets. nuScenes covers complex urban roads and intersections with rich multimodal context, while highD provides high-frequency highway trajectories with smooth and continuous motion. We refine the map processing module by reconstructing lane structures from highD and rasterizing them into local grid maps centered on the ego vehicle. This ensures that generated scenarios remain contextually consistent and adaptable to diverse road geometries. The combined dataset usage enables the model to generalize across traffic domains, covering safety-critical conditions from urban intersections to high-speed highway interactions.

\subsubsection{CVAE Model Architecture}

As the core trajectory generation model, the CVAE is responsible for learning a low-dimensional latent space, $Z$, from vehicle trajectory data and their interactions within the traffic data. It then decodes this latent space $Z$ to generate target trajectory scenarios. The architecture consists of three main components:

\textbf{Encoder Network} $q_{\phi}(z \mid X, \mathcal{M})$:  
The encoder maps high-dimensional input data, namely the historical trajectories $X$ of all agents and the map information $\mathcal{M}$, into the latent space $\mathcal{Z}$. Specifically, it estimates the parameters of the posterior Gaussian distribution of the latent variable $z$, including the mean $\mu_{z}$ and variance $\sigma_{z}^{2}$:
\begin{equation}
\mu_{z},\ \log \sigma_{z}^{2} = \text{Encoder}_{\phi}(X, c),
\end{equation}
where $\text{Encoder}_{\phi}$ is the encoder network parameterized by $\phi$, $c$ is the context vector derived from $\mathcal{M}$, and $(\mu_{z}, \sigma_{z}^{2})$ parameterize the approximate posterior distribution $q_{\phi}(z \mid X, \mathcal{M})$.  

\textbf{Decoder Network} $p_{\theta}(Y \mid z, \mathcal{M})$:  
The decoder is responsible for trajectory generation. A latent vector $z \sim \mathcal{N}(\mu_{z}, \sigma_{z}^{2})$ is sampled and, conditioned on $z$ and map features $\mathcal{M}$, the decoder produces joint future trajectories $\hat{Y}$ for all agents:
\begin{equation}
\hat{Y} = \text{Decoder}_{\theta}(z, c),
\end{equation}
where $\text{Decoder}_{\theta}$ is the decoder network parameterized by $\theta$, and $\hat{Y}$ denotes the predicted future trajectories conditioned on $z$ and $c$.  

\textbf{Prior Network} $p_{\psi}(z \mid \mathcal{M})$:  
The prior network estimates the prior distribution of the latent variable $z$ based solely on the map context $\mathcal{M}$. During training, this network serves as a regularization term to align the posterior distribution with the learned prior:
\begin{equation}
\mu_{\text{prior}},\ \sigma_{\text{prior}} = \text{Prior}_{\psi}(c),
\end{equation}
where $\text{Prior}_{\psi}$ is the prior network parameterized by $\psi$, and $(\mu_{\text{prior}}, \sigma_{\text{prior}})$ define the prior distribution.  

\textbf{Loss Function} $\mathcal{L}_{\text{CVAE}}$:  
The overall loss function consists of a reconstruction term, a Kullback–Leibler (KL) divergence term, and an interaction-based collision term:
\begin{equation}
\mathcal{L}_{\text{CVAE}} = \mathcal{L}_{\text{recon}} + w_{\text{KL}} \mathcal{L}_{\text{KL}} + w_{\text{coll}} \mathcal{L}_{\text{coll}},
\label{eq:cvae_total_loss}
\end{equation}
where $w_{\text{KL}}$ and $w_{\text{coll}}$ are weighting coefficients. The individual loss components are defined as:
\begin{align}
\mathcal{L}_{\text{recon}} &= -\mathbb{E}_{q_{\phi}(z \mid X, c)} \big[ \log p_{\theta}(X \mid z, c) \big], \label{eq:recon_loss}\\
\mathcal{L}_{\text{KL}} &= D_{\text{KL}} \big( q_{\phi}(z \mid X, c) \, \| \, p(z \mid c) \big), \label{eq:kl_loss}\\
\mathcal{L}_{\text{coll}} &= f_{\text{collision}}(\hat{Y}), \label{eq:collision_loss}
\end{align}
where $\mathcal{L}_{\text{recon}}$ measures reconstruction fidelity, $\mathcal{L}_{\text{KL}}$ enforces regularization of the latent space by aligning the posterior with the prior, and $\mathcal{L}_{\text{coll}}$ penalizes physically infeasible outcomes such as collisions between agents. Together, these terms ensure that the CVAE generates scenarios that are both realistic and risk-sensitive.

\subsubsection{Modeling Multi-Agent Interaction Based on GNNs}
Following the STRIVE framework, to capture the relational dynamics in traffic scenarios, we integrated the GNN into our model architecture to precisely model the dynamic interactions between agents. 

\textbf{Scene Graph Construction}:  
At each time step along the historical trajectory, the scene is represented as a directed graph $\mathcal{G} = (\mathcal{V}, \mathcal{E})$. Each vehicle is modeled as a node $v_i \in \mathcal{V}$, with an associated feature vector $h_i^{(0)}$ that encodes its historical kinematic state, including position, velocity, and heading. Interactions between vehicles $i$ and $j$ are represented by edges $e_{ij} \in \mathcal{E}$, whose features describe relative attributes such as distance, velocity difference, and potential conflict measures. This graph-based formulation allows the model to incorporate structured information about the environment and the relationships among agents.  

\textbf{Message Passing Mechanism}:  
The evolution of vehicle interactions is modeled by iteratively updating node representations through message passing. At each layer $l$, the hidden state of node $i$, denoted $h_i^{(l)}$, aggregates information from its neighbors $\mathcal{N}(i)$:  
\begin{equation}
h_i^{(l+1)} = \text{Update}\Bigg( h_i^{(l)},\ \bigoplus_{j \in \mathcal{N}(i)} \text{Message}\big(h_i^{(l)}, h_j^{(l)}, e_{ij}\big) \Bigg),
\end{equation}
where $\text{Message}(\cdot)$ defines the transformation applied to information from a neighboring node $j$, $\bigoplus$ denotes an aggregation operator such as summation or averaging, and $\text{Update}(\cdot)$ is typically a multilayer perceptron that refines the state of node $i$. By stacking multiple layers, the GNN captures both first-order interactions and higher-order dependencies that emerge through indirect connections.


\subsection{Knowledge-Guided Loss Function Optimization with LLMs}

\begin{figure}[htbp]
    \centering
    \includegraphics[width=1.0\textwidth]{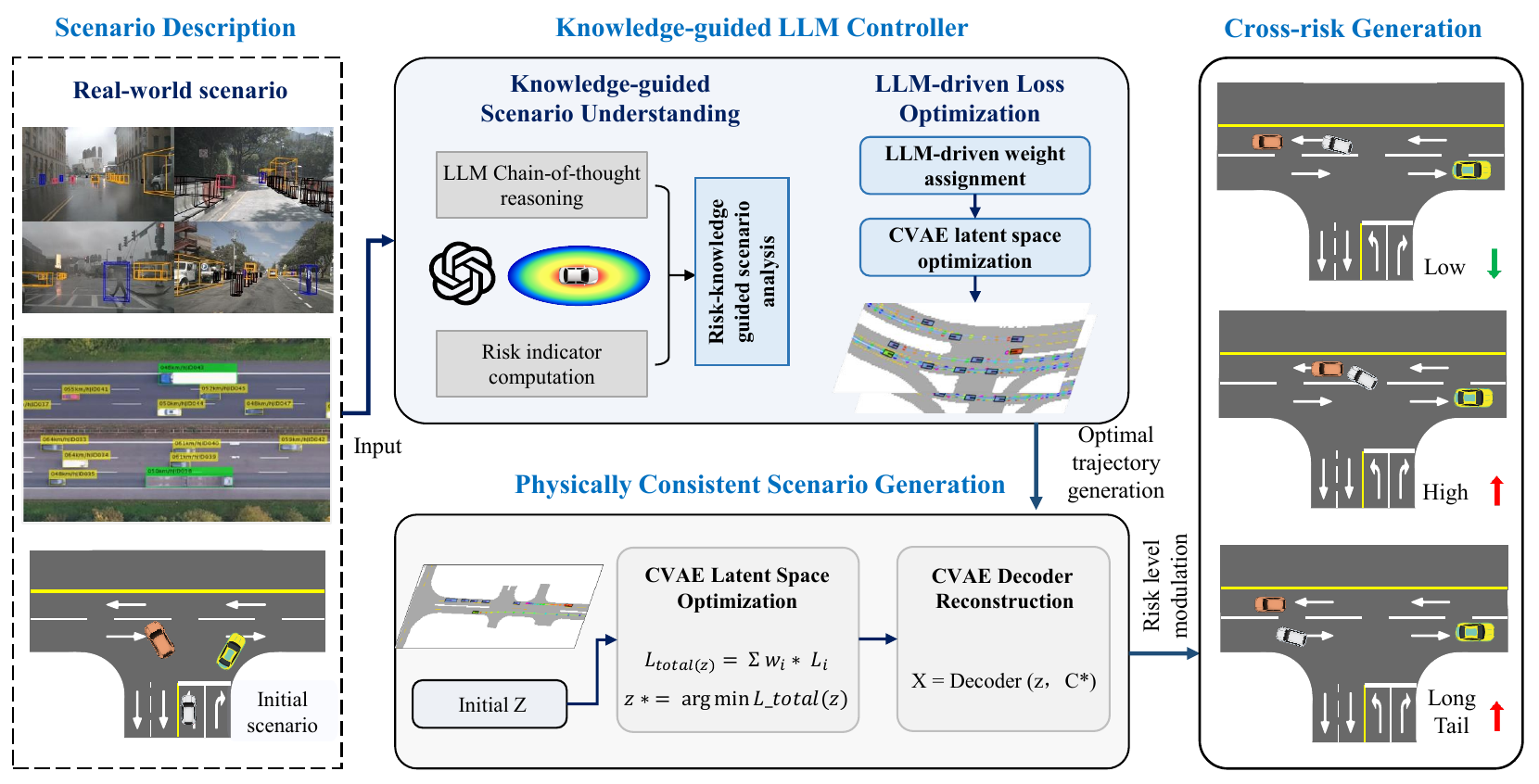}
    \caption{Overview of the knowledge-guided loss function optimization framework. 
The LLM interprets scene semantics and risk indicators through chain-of-thought reasoning, computes adaptive loss weights, and guides CVAE latent-space optimization for generating physically consistent and risk-controllable traffic scenarios.}
    \label{fig:framework_loss}
\end{figure}

Within the CVAE framework, the generation of realistic trajectories depends on effectively shaping the latent space distribution \( Z \). The form and weighting of loss terms determine this distribution, influencing the realism and risk properties of generated behaviors. To achieve controllable and interpretable risk-aware scenario generation, the loss function must be dynamically generated and optimized according to contextual risk knowledge rather than using fixed weights.
We propose a knowledge-guided reasoning framework powered by a LLM, which generates and refines loss functions to adaptively guide CVAE optimization. As shown in Fig.~\ref{fig:framework_loss}, the framework forms a closed loop between scenario understanding, semantic reasoning, and latent-space optimization. It consists of two components: a knowledge-guided scenario understanding module that extracts structured risk cues from unstructured inputs, and an LLM-based loss generation and optimization module that transforms these cues into quantitative loss terms and adaptive weighting coefficients.
The LLM interprets scene semantics, analyzes multi-agent interactions, and computes risk indicators such as TTC, minimum lateral distance, and curvature. Based on these cues, it synthesizes a set of weighted sub-loss functions and adjusts their coefficients \( w_i \) to match the detected risk context. The total optimization objective is expressed as
\begin{equation}
L_{\text{total}}(z) = \sum_i w_i L_i(\text{Decoder}(z, C)),
\quad
z^{*} = \arg\min_{z} L_{\text{total}}(z),
\label{eq:total_loss}
\end{equation}
where the generated loss terms ensure convergence toward physically feasible yet risk-sensitive trajectories. Through iterative reasoning and gradient feedback, the LLM continuously refines both the structure and weights of the loss functions. The decoder then reconstructs trajectories \( \hat{Y} = \text{Decoder}(z^{*}, C) \) that are physically consistent and span low-, high-, and long-tail risk regimes, forming an end-to-end knowledge-guided optimization process within the CVAE-GNN framework.

\subsubsection{Traffic Scenario Description and Analysis}

Using only a CVAE or a fixed loss function makes it difficult to balance both ``realism'' and a ``controllable risk distribution.'' To achieve this, the LLM performs structured reasoning on each scenario, transforming heterogeneous inputs into quantitative cues that guide loss adaptation within the CVAE optimization process. The process includes two stages: structured scenario description and LLM-based reasoning and evaluation.

\textbf{Scenario description inputs.}
(1) Dynamic trajectories: vehicle positions, velocities, and headings over time, converted from numerical coordinates into textual descriptions for interpretability.  
(2) Static map context: road geometry, lane boundaries, traffic signals, and obstacles defining the physical environment.  
(3) Relational cues: inter-vehicle distances, relative speeds, and potential conflict zones that characterize multi-agent interactions.  

\textbf{LLM-based reasoning and evaluation.}
(1) Behavior knowledge base: a library of predefined behaviors (e.g., car-following, lane change, sudden braking) covering both low- and high-risk cases, constructed from historical scenarios and multimodal sources.  
(2) Chain-of-thought reasoning: a structured multi-step analysis that decomposes the scenario into agent behaviors, spatial–temporal relations, and risk indicators such as TTC, minimum distance, and clearance, identifying interactions likely to evolve into long-tail risks.  
(3) Structured outputs: the LLM expresses reasoning results in machine-readable form, e.g.,  
\texttt{\{"side": "right", "type": "cut\_in", "strength": 0.9\}}, which are passed to downstream modules such as the field-based risk model.  
(4) Self-evaluation: each reasoning chain is verified for logical completeness and consistency, and a confidence score is assigned to indicate reliability.  

This layered process converts unstructured scenario descriptions into structured, optimizable representations, allowing the LLM to dynamically guide loss weighting and improve the interpretability and controllability of risk-aware scenario generation.

\subsubsection{Risk Indicator Design}

Following the semantic reasoning in Section~3.3.1, the next step is to transform high-level behavioral intents into quantitative objectives that can be optimized in the CVAE latent space. This requires identifying where risk is most likely to occur and quantifying how severe the interaction may be. To this end, we design a risk indicator module that integrates a potential field model with auxiliary indicator computations.

\textbf{Potential field model.}  
The potential field model for risk point computation translates semantic intent (e.g., cut-in, sudden braking) into a concrete geometric anchor. The driving environment is represented as a two-dimensional scalar field, where each point $p=(x,y)$ has potential $U(p)$. The total potential is
\begin{equation}
  U_{\text{total}}(p) = w_{\text{rep}} U_{\text{rep}}(p) + w_{\text{attr}} U_{\text{attr}}(p),
\end{equation}
where $U_{\text{rep}}$ encodes collision and boundary avoidance, $U_{\text{attr}}$ attracts the ego vehicle toward intent-consistent regions, and $w_{\text{rep}}, w_{\text{attr}}>0$ balance their influence. The equilibrium $p^{*}=\arg\min_{p} U_{\text{total}}(p)$ denotes the most critical point for a potential high-risk event.

\textbf{Auxiliary indicators.}  
Although $p^*$ localizes the risk, it does not fully describe the interaction intensity. We therefore extend it into a target vector
\begin{equation}
\mathbf{T} = \big[\,p^*,\; d_{\text{lat}}^*,\; \text{TTC}^*,\; \kappa^*\,\big],
\label{eq:target_vector}
\end{equation}
where (1) \emph{Minimum lateral gap $d_{\text{lat}}^*$.} A small lateral distance is a strong indicator of cut-ins or side collisions. By explicitly constraining $d_{\text{lat}}^*$, we ensure that generated scenarios can capture close-proximity interactions, which are common triggers for evasive maneuvers.
(2) \emph{Time-to-collision $\text{TTC}^*$.} TTC directly reflects the urgency of the interaction. Very small values (e.g., $\leq 1$s) indicate critical, near-miss situations that are rarely observed in real data but crucial for safety testing. Explicitly targeting $\text{TTC}^*$ allows the framework to generate long-tail cases that challenge ADS response times.  
(3) \emph{Path curvature $\kappa^*$.} This describes the geometric difficulty of the maneuver, for example sharp lane changes or turning trajectories. Incorporating curvature ensures that scenarios remain both dynamically realistic and physically consistent.

By combining field-based localization with these auxiliary indicators, semantic intents are grounded into a structured and interpretable target vector $\mathbf{T}$. This representation captures both \emph{where} and \emph{how} a risky interaction occurs, and serves as the direct input for subsequent loss weight generation, linking high-level reasoning with trajectory-level optimization.

\subsubsection{LLM-Driven Long-Tail Behavior Generation and Loss Optimization}

To generate long-tail, safety-critical scenarios, it is not sufficient to localize risk points and compute indicators; these semantic cues must be translated into optimization objectives that directly shape trajectory synthesis. In our framework, this role is fulfilled by the LLM, which serves as a high-level controller. Given the semantic description and the indicator vector $\mathbf{T}$, the LLM assigns weights to different loss terms, thereby determining how strongly each behavioral constraint should influence latent-space optimization. For example, when $\text{TTC}^*$ is extremely small, the weight of the TTC penalty is increased to enforce urgent interactions, while a narrow $d_{\text{lat}}^*$ raises the weight on lateral-distance penalties to induce close-proximity cut-ins. In this way, abstract semantic risks are translated into concrete optimization signals that bias trajectory generation toward high-risk regions.

To improve consistency and interpretability, the LLM is provided with structured definitions of available loss functions and a compact corpus of representative \{scenario, weight\} exemplars. This enables few-shot reasoning, allowing the model to generalize from canonical traffic behaviors (e.g., overtaking, sudden braking) to novel risk combinations. The output is a weight vector $\mathbf{w}$ that balances adversarial challenge with physical plausibility. Through this mechanism, unstructured scene descriptions are systematically transformed into actionable optimization objectives, bridging high-level reasoning and low-level trajectory generation.

In the CVAE-based generator, a latent variable $\mathbf{z}\in\mathcal{Z}$ parameterizes a trajectory $\mathbf{X}'\in\mathcal{X}$ through the decoder:
\begin{equation}
\mathbf{X}' = \mathrm{Decoder}(\mathbf{z}, \mathbf{C}),
\label{eq:decoder}
\end{equation}
where $\mathbf{C}$ denotes contextual information such as road geometry or surrounding vehicles. This mapping is differentiable, ensuring that changes in $\mathbf{z}$ propagate to the generated trajectory.

The total objective is a weighted sum of multiple design terms:
\begin{equation}
\mathcal{L}_{\mathrm{total}}(\mathbf{z}) = \sum_{i=1}^{n} w_i\, \mathcal{L}_i\!\left(\mathrm{Decoder}(\mathbf{z}, \mathbf{C})\right),
\label{eq:ltotal}
\end{equation}
where each $\mathcal{L}_i$ encodes a property such as collision proximity, time-to-collision, or kinematic smoothness, and $w_i$ is the weight assigned by the LLM. The optimization seeks
\[
\mathbf{z}^*=\arg\min_{\mathbf{z}} \mathcal{L}_{\mathrm{total}}(\mathbf{z}),
\]
ensuring trajectories respect both realism and the desired risk level.

When a new behavioral prior is introduced, the objective becomes:
\begin{equation}
\mathcal{L}'_{\mathrm{total}}(\mathbf{z})=\mathcal{L}_{\mathrm{total}}(\mathbf{z})+w_{\mathrm{new}}\,\mathcal{L}_{\mathrm{new}}\!\left(\mathrm{Decoder}(\mathbf{z}, \mathbf{C})\right).
\label{eq:ltotal_new}
\end{equation}
This modifies the gradient field that guides optimization:
\begin{equation}
\nabla_{\mathbf{z}}\mathcal{L}'_{\mathrm{total}}(\mathbf{z})=\nabla_{\mathbf{z}}\mathcal{L}_{\mathrm{total}}(\mathbf{z})+w_{\mathrm{new}}\nabla_{\mathbf{z}}\mathcal{L}_{\mathrm{new}}(\mathbf{z}),
\label{eq:grad_total}
\end{equation}
where the additional term is computed by the chain rule:
\begin{equation}
\nabla_{\mathbf{z}}\mathcal{L}_{\mathrm{new}}(\mathbf{z})
=\frac{\partial \mathcal{L}_{\mathrm{new}}}{\partial \mathbf{X}'}\,
\frac{\partial \mathbf{X}'}{\partial \mathbf{z}}, \quad \mathbf{X}'=\mathrm{Decoder}(\mathbf{z},\mathbf{C}).
\label{eq:chain_rule}
\end{equation}

Finally, under a learning rate $\eta>0$, the latent update becomes:
\begin{equation}
\mathbf{z}_{t+1}=\mathbf{z}_t-\eta\left(
\nabla_{\mathbf{z}}\mathcal{L}_{\mathrm{total}}(\mathbf{z}_t)
+w_{\mathrm{new}}\,\nabla_{\mathbf{z}}\mathcal{L}_{\mathrm{new}}(\mathbf{z}_t)\right).
\label{eq:update}
\end{equation}

Equation~\eqref{eq:update} shows that new loss terms, weighted by LLM-derived coefficients, bias the optimization path toward regions that satisfy the additional behavioral prior. The system therefore converges to a new equilibrium $\mathbf{z}'^*$ that balances the original objectives with risk-sensitive requirements.
This design establishes a direct pipeline from semantic risk descriptions to differentiable optimization. By combining LLM-driven weight assignment with latent-space gradient updates, the framework systematically generates adversarial yet feasible trajectories across a continuum of risk levels.

\subsection{Composite Objective for Adversarial Yet Feasible Trajectories}

We target trajectories that are adversarial for testing, physically consistent, and geometrically compatible with the environment. We decompose the objective into three groups,
\begin{equation}
\mathcal{L}_{\mathrm{total}}=\alpha_{1}\,\mathcal{L}_{\mathrm{traj}}+\alpha_{2}\,\mathcal{L}_{\mathrm{geom}}+\alpha_{3}\,\mathcal{L}_{\mathrm{interact}},
\label{eq:composite_total}
\end{equation}
where $\mathcal{L}_{\mathrm{traj}}$ captures single-agent kinematics, $\mathcal{L}_{\mathrm{geom}}$ enforces map and geometry consistency, and $\mathcal{L}_{\mathrm{interact}}$ encodes multi-agent interaction and adversarial intent. the coefficients $\alpha_i$ are set by a structured risk assessment that reflects safety-critical event criteria, not by ad hoc tuning.

\subsubsection{Collision-Aware Physical Constraints}

\textbf{Point-to-point collision loss}: We pull the ego front toward the victim rear to induce rear-end contact:
\begin{equation}
\mathcal{L}_{\mathrm{collision\text{-}point}}=\left\|\mathbf{p}^{\mathrm{front}}_{\mathrm{ego}}-\mathbf{p}^{\mathrm{rear}}_{\mathrm{victim}}\right\|_{2}.
\label{eq:collision_point}
\end{equation}
Here $\mathbf{p}^{\mathrm{front}}_{\mathrm{ego}},\mathbf{p}^{\mathrm{rear}}_{\mathrm{victim}}\in\mathbb{R}^{2}$ or $\mathbb{R}^{3}$ denote bumper center positions. the resulting gradient provides a strong, continuous pull toward physical contact. this loss is effective when combined with orientation or speed control terms that shape the approach profile.

\textbf{Side-impact minimum-distance loss}:
To elicit lateral impacts, we attract the ego front toward the closest point on the victim body:
\begin{equation}
\mathcal{L}_{\mathrm{side\text{-}impact}}
=\min_{\mathbf{p}\in\mathcal{B}_{\mathrm{victim}}}
\left\|\mathbf{p}^{\mathrm{front}}_{\mathrm{ego}}-\mathbf{p}\right\|_{2},
\label{eq:side_impact}
\end{equation}
where $\mathcal{B}_{\mathrm{victim}}$ is a bounding-box proxy of the victim geometry. the inner minimization updates the target point as geometry evolves, producing a dynamic target-seeking behavior that is suitable for T-bone and aggressive cut-in scenarios.

\subsubsection{Behavioral Consistency and Interaction Objectives}

\textbf{Time-to-collision penalty}:
We penalize unsafe approach rates whenever estimated TTC falls below a threshold:
\begin{equation}
\mathcal{L}_{\mathrm{TTC}}=\max\!\left(0,\;T_{\mathrm{safe}}-\mathrm{TTC}\right)^{2}.
\label{eq:ttc}
\end{equation}
here $T_{\mathrm{safe}}$ is a safety margin, for example 2 seconds, and TTC is computed from relative speed along the line of sight. the quadratic penalty strengthens gradients near the critical boundary, which helps the optimizer cross from safe to near-collision regimes during adversarial search.

\textbf{Lateral gap shaping}:
We reduce the free lateral clearance between vehicles to provoke challenging interactions:
\begin{equation}
\mathcal{L}_{\mathrm{lat}}=k\,(d_{\mathrm{lat\text{-}gap}})^{2},
\label{eq:lateral}
\end{equation}
with
\begin{equation}
d_{\mathrm{lat\text{-}gap}}=d_{\mathrm{lat}}-\left(\frac{w_{\mathrm{ego}}}{2}+\frac{w_{\mathrm{target}}}{2}\right).
\label{eq:lat_gap}
\end{equation}
where $d_{\mathrm{lat}}$ is the center-to-center lateral distance, $w_{\mathrm{ego}}$ and $w_{\mathrm{target}}$ are vehicle widths, and $k>0$ is a weight. the smooth quadratic form supports stable gradients. as $d_{\mathrm{lat\text{-}gap}}\to 0$, edges approach contact, and negative values indicate overlap, which corresponds to collision.

\textbf{Yaw-rate aggressiveness}:
We encourage sharp, uncomfortable steering to stress controllers:
\begin{equation}
\mathcal{L}_{\mathrm{yaw}}=\frac{1}{\max\!\left(0,\,|\omega|-\omega_{\mathrm{thresh}}\right)+\epsilon},
\label{eq:yaw}
\end{equation}
where $\omega$ is the planned yaw rate, $\omega_{\mathrm{thresh}}$ is a comfort threshold, and $\epsilon>0$ avoids division by zero. The inverse form yields large penalties when $|\omega|\le\omega_{\mathrm{thresh}}$, and it decays as $|\omega|$ exceeds the threshold, which incentivizes high-aggression maneuvers during adversarial generation.

Equations~\eqref{eq:collision_point}–\eqref{eq:yaw} shape complementary aspects of interaction, proximity, and maneuver aggressiveness. combined within \eqref{eq:composite_total}, they expose clear, differentiable pathways for gradients to flow from high-level behavioral priors to latent updates via \eqref{eq:chain_rule} and \eqref{eq:update}. this design enables targeted, reproducible generation of physically feasible yet safety-critical scenarios for evaluation.

\subsection{Cross-Risk Scenario Generation and Distribution Control}

The optimized CVAE-LLM framework enables not only realistic trajectory generation but also controllable risk modulation across diverse traffic conditions. To ensure that generated scenarios systematically cover the full spectrum of driving risk, from normal to extreme long-tail events, we establish a unified risk definition and dynamic distribution control strategy.

\textbf{Risk level definition:}
To comprehensively evaluate autonomous driving safety, scenarios are categorized into three levels: low-risk, high-risk, and long-tail, as summarized in Table~\ref{tab:risk_def}. These represent a continuous spectrum from stable, everyday traffic to rare but safety-critical events.

\begin{table}[htbp]  
\centering  
\caption{Definitions and examples of scenario risk levels}  
\label{tab:risk_def}
\begin{tabularx}{\textwidth}{l X X}  
\toprule  
\textbf{Level} & \textbf{Key Characteristics} & \textbf{Examples} \\
\midrule  
Low-risk & Stable flow, sufficient margins, minimal interaction. & Car-following, cruising, compliant lane change. \\
High-risk & Strong interactions, short reaction time, safety-critical gaps. & Sudden braking, dense cut-ins, occluded crossing. \\
Long-tail & Rare, extreme, or adversarial events. & Very short TTC, skidding, high-speed violations. \\
\bottomrule  
\end{tabularx}  
\end{table}

\textbf{Dynamic risk control:}
Data-driven generators tend to reproduce low-risk distributions, which are inadequate for evaluating safety boundaries. To address this limitation, we design a closed-loop control process that dynamically balances the proportion of scenarios across different risk levels, as summarized in Table~\ref{tab:risklevel}. Quantitative indicators are used for risk assessment, including: (1) TTC (time to collision), representing the remaining time before two agents collide under current speeds; (2) THW (time headway), the temporal distance to the preceding vehicle; and (3) TLC (time to lane crossing), measuring lateral stability under current motion.  

The field model continuously monitors these indicators to identify risk transitions and guide the sampling process. When risk indicators fall below predefined thresholds (see Table~\ref{tab:risklevel}), the model triggers higher-risk scenario generation by adjusting latent sampling and loss weights. This mechanism ensures that the generated scenarios maintain both diversity and controllable coverage across low-, high-, and long-tail conditions. Consequently, the risk-driven control mechanism enables the framework to generate balanced, physically consistent, and safety-critical traffic scenarios, supporting comprehensive validation of autonomous driving systems (ADS) under varying levels of interaction intensity and risk exposure.

\begin{table}[htbp]  
    \centering  
    \small
    \caption{Quantitative thresholds for scenario risk levels}  
    \label{tab:risklevel}  
    \begin{tabular}{l c c c}  
        \toprule  
        \textbf{Risk Level} & \textbf{TTC (s)} & \textbf{TLC (s)} & \textbf{THW (s)} \\
        \midrule  
        Low Risk & $\geq$ 5–7 & $\geq$ 1.5–2.0 & $\geq$ 2.5–3.0 \\  
        \addlinespace  
        High Risk & 1.5–3.0 & 0.8–1.5 & 1.0–2.5 \\  
        \addlinespace  
        Long-tail Events & $\leq$ 1.5 (esp. $\leq$ 1.0) & $\leq$ 0.8 (esp. $\leq$ 0.5–0.6) & $\leq$ 1.0 (esp. $\leq$ 0.5–0.7) \\  
        \bottomrule  
    \end{tabular}  
\end{table}

\section{Experiment and Results Analysis}

This section presents the experimental evaluation of the proposed framework and its effectiveness in generating safety-critical scenarios. We first evaluate the CVAE module to verify its ability to produce realistic and diverse foundational scenarios as the basis for subsequent optimization. Then, we assess how LLM-driven reasoning enhances adversarial generation across different risk levels, particularly in covering long-tail events. Accordingly, the experiments focus on three aspects: (1) the realism and diversity of CVAE-generated scenarios across multiple datasets; (2) the contribution of LLM-guided optimization to risk-sensitive generation; and (3) the overall improvement in long-tail coverage and system robustness.

\subsection{Datasets and Preprocessing}

Our experiments are conducted on two large-scale naturalistic driving datasets: highD and nuScenes, which together cover diverse traffic environments from structured highways to complex urban intersections.  

\begin{itemize}
    \item \textbf{highD.} Collected on German highways using drones, highD provides high-frequency vehicle trajectories that capture lane changes, merges, and other high-risk maneuvers in structured traffic. Static aerial images are converted into two layers (\texttt{drivable\_area} and \texttt{lane\_divider}) and rasterized into $420 \times 100$ meter local maps at a resolution of 2 pixels per meter, centered on the ego vehicle.  

    \item \textbf{nuScenes.} Recorded in the cities of Singapore and Boston, nuScenes offers multimodal sensor data, dynamic object annotations, and detailed HD maps covering intersections, curved roads, and occluded urban areas. Vectorized map elements such as lane dividers, stop lines, and traffic signals are directly incorporated into the preprocessing pipeline to preserve spatial and semantic consistency.  
\end{itemize}

This unified preprocessing enables the CVAE to learn a shared latent representation that captures both structured highway motion and complex urban interactions, providing a diverse foundation for risk-aware scenario generation and distribution control.

\subsection{Simulation Environments and Baselines}

To evaluate the effectiveness of the generated scenarios in a closed-loop setting, we conduct experiments on two mainstream simulation platforms that complement each other in focus and fidelity:
\begin{itemize}
    \item \textbf{CARLA:} A high-fidelity simulator providing realistic physics and visual rendering, used to assess the perception and decision-making performance of an autonomous driving system (ADS) under near-real-world conditions.
    \item \textbf{SMARTS:} A scalable multi-agent simulator emphasizing interactive traffic flow and behavioral diversity, suitable for testing the strategic reasoning and cooperation capabilities of an ADS in dense, dynamic environments.
\end{itemize}

Furthermore, two baseline settings are established for comparison to highlight the framework’s advantages.

\begin{itemize}
    \item \textbf{Base Scenarios:} Scenarios generated directly by the CVAE without LLM guidance. This serves as the main baseline for quantifying the incremental contribution of the LLM. The baseline integrates a VAE with a GNN that encodes historical trajectories and contextual scenario information to generate future trajectory sequences in an autoregressive manner. The generation process is constrained by a composite loss function with fixed empirical weight coefficients ($w_1, w_2, w_3$), preventing adaptive adjustment to varying risk levels. As a purely data-driven model, it may produce statistically plausible but physically infeasible trajectories, and its generalization to rare “long-tail” scenarios such as emergency evasions remains limited.  

    \item\textbf{Parametric Perturbation:} A baseline that introduces random perturbations to key motion parameters (e.g., vehicle acceleration, lane-change duration) in the base scenarios. Although this method provides limited variability through parameter noise, it remains scenario-independent and lacks semantic understanding, illustrating the superiority of the LLM-guided adaptive optimization over naive random perturbations.
\end{itemize}

\subsection{Evaluation Metrics}

To comprehensively and quantitatively evaluate the proposed two-stage scenario generation framework, we define key metrics across four dimensions: trajectory realism, interaction consistency, long-tail event coverage, and robustness against physical plausibility violations.  

\textbf{Trajectory Realism}: Trajectory realism measures the similarity between CVAE-generated trajectories and real-world trajectories. We define a composite realism score $TR$ as the weighted average of multiple sub-metrics:  
\begin{equation}
TR = \frac{1}{n} \sum_{i=1}^{n} M_i, \quad M_i = Z_m + Z_t + Z_s + Z_{ms} + Z_{tc},
\end{equation}
where $n$ is the number of sub-metrics. $Z_m$ denotes the deviation distance in latent space between adversarial and original samples, $Z_t$ represents the standard deviation of the adversarial latent distribution, $Z_s$ is the log-likelihood score of adversarial variables under the prior, $Z_{ms}$ measures kinematic smoothness (e.g., jerk), and $Z_{tc}$ quantifies geometric continuity of trajectories.  

\textbf{Interaction Consistency}: Interaction consistency ensures that vehicle interactions in adversarial scenarios remain physically plausible and logically consistent. We define a composite interaction score $I$ as:  
\begin{equation}
I = \frac{1}{5}(D_r + V_c + P_r + S_c + C_r),
\end{equation}
where $D_r$ denotes inter-vehicle distance plausibility, $V_c$ captures velocity continuity, $P_r$ reflects probabilistic reasoning of feasible maneuvers, $S_c$ represents spatial coherence among interacting vehicles, and $C_r$ accounts for collision risk balance. For example, $D_r$ ensures that inter-vehicle distances remain within reasonable ranges, avoiding unrealistic overlap or excessive separation.  

\textbf{Long-Tail Event Coverage}: To quantify the ability of generated scenes to cover long-tail events, we evaluate the scenes based on a predefined metric system. A generated scene is identified as a long-tail event if and only if it simultaneously satisfies the threshold requirements for all three "long-tail" related metrics defined in Table~\ref{tab:risklevel}. We define Long-Tail Coverage Rate (LCR) as the proportion of scenarios that satisfy the thresholds of three long-tail metrics. Specifically,
\begin{equation}
    \mathrm{LCR} \;=\; \frac{ \left| \left\{\, s \in \mathcal{S} \;\middle|\; I_1(s) \ge \tau_1,\; I_2(s) \ge \tau_2,\; I_3(s) \ge \tau_3 \,\right\} \right| }{ \left| \mathcal{S} \right| },
\end{equation}
where \(I_1, I_2, I_3\) are the metric functions described in Table 2, \(\tau_1, \tau_2, \tau_3\) are their respective thresholds, and \(|\cdot|\) denotes the cardinality of a set.

\textbf{Sim-to-Real}: This metric aims to quantify the discrepancy between generated trajectories and real-world data, ensuring the authenticity of the generated scenarios.We assess the Sim-to-Real gap across four dimensions: statistical distribution, motion features, frequency domain characteristics, and high-level semantic behaviors
\begin{equation}
    \mathrm{STR} = \frac{1}{4} \left( G_{\text{dist}} + G_{\text{motion}} + G_{\text{freq}} + G_{\text{sem}} \right),
\end{equation}
where \(G_{\text{dist}}\) calculates the difference in the distribution of key kinematic features between simulated trajectories (\( \mathcal{T}_{\text{sim}} \)) and real trajectories (\( \mathcal{T}_{\text{real}} \)).\(G_{\text{motion}}\) measure the discrepancy in basic kinematic statistics.\(G_{\text{freq}}\) evaluate the smoothness and periodicity of motion by comparing the characteristics of trajectories in the frequency domain.\(G_{\text{sem}}\) assess the frequency differences in high-level driving behavior patterns.

\subsection{Validation of Base Scenario Generation}

We first evaluate the CVAE module before introducing any adversarial refinement. As shown in Table~\ref{tab:base_cvae}, the CVAE achieves a mean squared error below $0.5\,\text{m}^2$ on both highD and nuScenes, indicating accurate spatial reconstruction. The KL divergence for acceleration distributions remains within $[1.08,\,2.75]$, and the curvature KL stays below $0.30$, confirming that the generated trajectories preserve the key statistical properties of real vehicle dynamics.

These results indicate that the CVAE-GNN learns a stable and structured latent space capable of reproducing diverse, physically consistent motion patterns across both highway and urban environments. Qualitative inspection further shows coherent lane-keeping, merging, and lane-changing behaviors, suggesting that the model captures fundamental multi-agent interaction patterns.
This high-quality generative prior is essential for the subsequent LLM-guided refinement. Because the latent space is already physically grounded and distributionally faithful, the adversarial optimization process remains stable and controllable, enabling the full framework to explore higher-risk regions without producing unrealistic artifacts.

\begin{table}[htbp]
\centering
\small
\caption{Quantitative Evaluation of CVAE Base Scenario Generation on highD and nuScenes}
\label{tab:base_cvae}
\begin{tabular}{lccc}
\toprule
\textbf{Metric} & \textbf{nuScenes} & \textbf{highD} & \textbf{Description} \\
\midrule
MSE ($\text{m}^2$) ↓ & 0.460 & 0.420 & Trajectory reconstruction error \\
Acceleration KL ↓ & 2.75 & 1.08 & KL divergence of acceleration distribution \\
Curvature KL ↓ & 0.17 & 0.26 & KL divergence of curvature distribution \\
\bottomrule
\end{tabular}
\end{table}

\subsection{Evaluation of LLM-Based Adversarial Enhancement}
\label{sec:llm_enhancement}

We evaluate the effectiveness of the LLM-based adversarial reasoning module, which acts as an intelligent engine to refine risk-sensitive scenarios and expose ADS to rare but critical interactions. A total of 120 simulation trials were conducted in CARLA, covering both traffic risk cases (TRC) and relatively safe conditions.

\textbf{Overall Performance.}  
Table~\ref{tab:main_results_modified} compares different scenario generation methods. The proposed CVAE + LLM framework achieves the highest long-tail event coverage of 22.8\%, exceeding both the CVAE-only baseline (17\%) and the CVAE + Perturbation variant (21.6\%). Meanwhile, trajectory realism (52.34\%), interaction consistency (75.4\%), and sim-to-real transferability (51.1\%) remain comparable to baseline levels. This demonstrates that the LLM-based reasoning effectively expands coverage of rare, high-risk events without sacrificing the realism or coherence of generated trajectories.

\begin{table}[htbp]
    \centering
    \small
    \begin{threeparttable}
        \caption{Performance Comparison of Different Scenario Generation Methods}
        \label{tab:main_results_modified}
        \begin{tabular}{l c c c c}
            \toprule
            \textbf{Method} &
            \makecell[c]{Long-Tail \\ Event Coverage (\%) \\ $\uparrow$} &
            \makecell[c]{Trajectory \\ Realism (\%) \\ $\uparrow$} &
            \makecell[c]{Interaction \\ Consistency (\%) \\ $\uparrow$} &
            \makecell[c]{Sim-to-Real (\%) \\ $\uparrow$} \\
            \midrule
            CVAE-Only (Base) & 17.0 & 53.9 & 75.5 & 49.8 \\
            CVAE + Perturb.  & 21.6 & 50.0 & 73.0 & 46.7 \\
            \textbf{CVAE + LLM (Ours)} & \textbf{22.8} & 52.3 & 75.4 & \textbf{51.1} \\
            \bottomrule
        \end{tabular}
    \end{threeparttable}
\end{table}

\begin{table}[htbp]
    \centering
    \small
    \renewcommand{\arraystretch}{1.25}
    \setlength{\tabcolsep}{8pt}
    \begin{threeparttable}
        \caption{Comparison of Long-Tail Event Generation Capability}
        \label{tab:long_tail_generation}
        \begin{tabularx}{0.95\textwidth}{@{} >{\centering\arraybackslash}m{3.8cm} 
                                            >{\centering\arraybackslash}m{3.6cm} 
                                            >{\centering\arraybackslash}m{3.6cm} 
                                            >{\centering\arraybackslash}m{3.6cm} @{}}
            \toprule
            \textbf{Method} &
            \textbf{Aggressive Cut-ins} &
            \textbf{Unsafe Merges} &
            \textbf{Dangerous Lane-change} \\
            \midrule
            CVAE-Only (Base) & 21.0 & 15.8 & 15.8 \\
            \textbf{CVAE + LLM (Ours)} & \textbf{35.9} & \textbf{20.5} & \textbf{25.6} \\
            \bottomrule
        \end{tabularx}
        \vspace{2pt}
        \begin{tablenotes}
            \footnotesize
            \item Note: Values indicate the proportion of generated scenarios identified as long-tail events. 
            Higher coverage (\%) indicates stronger capability to synthesize rare, safety-critical interactions.
        \end{tablenotes}
    \end{threeparttable}
\end{table}

\textbf{Long-tail Event Diversity.}  
As shown in Table~\ref{tab:long_tail_generation}, the LLM-based refinement substantially increases the coverage of high-risk and rare interactions. Specifically, the proportion of aggressive cut-ins rises from 21.0\% to 35.9\%, unsafe merges from 15.8\% to 20.5\%, and dangerous lane changes from 15.8\% to 25.6\%. These results indicate a broadening of the behavioral spectrum, covering critical driving patterns that are rarely captured by baseline stochastic sampling. The increase of approximately 40\% in long-tail coverage confirms that the adversarial reasoning mechanism effectively drives the generation toward challenging and safety-critical subspaces.

\textbf{Adversarial Impact on Scenario Risk.}  
Table~\ref{tab:main_results_2} compares the key performance indicators between the LLM-enhanced extremely challenging scenarios (TRC) and the relatively safe baseline scenarios.  
The collision rate reaches 97.1\% in the LLM-enhanced group, compared to 96.2\% in the baseline, confirming the higher frequency of critical interactions. The average TTC decreases drastically from 0.89\,s to 0.34\,s, while the average distance between vehicles shrinks from 3.03\,m to 0.30\,m. Similarly, the lateral TLC drops from 7.05\,s to 4.91\,s, indicating intensified lateral conflict dynamics. Although the average THW increases from 1.72\,s to 3.84\,s, this reflects temporal fluctuation in spacing control due to high-frequency evasive maneuvers.  
These metrics collectively validate that LLM-guided reasoning successfully amplifies adversarial risk exposure while maintaining physical realism.

\begin{table}[htbp]
    \centering
    \small
    \renewcommand{\arraystretch}{1.2}
    \setlength{\tabcolsep}{8pt}
    \begin{threeparttable}
        \caption{Comparison of Key Performance Indicators Between LLM-Enhanced and Baseline Scenarios}
        \label{tab:main_results_2}
        \begin{tabularx}{0.95\textwidth}{@{} >{\centering\arraybackslash}m{4cm} 
                                            >{\centering\arraybackslash}m{5cm} 
                                            >{\centering\arraybackslash}m{5cm} @{}}
            \toprule
            \textbf{Performance Indicator} & 
            \textbf{Extremely Challenging (LLM-Enhanced, TRC)} & 
            \textbf{Relatively Safe (Baseline Response)} \\
            \midrule
            Collision Rate (\%) $\uparrow$ & 97.1 & 96.2 \textcolor{red}{(-0.9)} \\
            Avg. TTC (s) $\downarrow$ & 0.34 & 0.89 \textcolor{blue}{(+0.55)} \\
            Avg. TLC (s) $\downarrow$ & 4.91 & 7.05 \textcolor{red}{(-2.14)} \\
            Avg. Distance (m) $\downarrow$ & 0.30 & 3.03 \textcolor{blue}{(+2.73)} \\
            Avg. THW (s) $\downarrow$ & 3.84 & 1.72 \textcolor{red}{(-2.12)} \\
            \bottomrule
        \end{tabularx}
        \vspace{2pt}
        \begin{tablenotes}
            \footnotesize
            \item Note: $\uparrow$ indicates higher values are better; $\downarrow$ indicates lower values are better. 
            Blue values denote increases, and red values denote decreases relative to the baseline.
        \end{tablenotes}
    \end{threeparttable}
\end{table}

Overall, the results show that the LLM-based adversarial reasoning significantly enhances the framework’s ability to discover rare yet consequential scenarios. Compared with conventional CVAE or random perturbation methods, the proposed approach increases long-tail event coverage by more than 30–40\%, broadens behavioral diversity, and exposes the ADS to intensified longitudinal and lateral conflicts (as evidenced by reduced TTC and inter-vehicle distances). Meanwhile, the realism and interaction consistency remain stable, confirming that the enhancement arises from intelligent adversarial reasoning rather than random disturbance. The generated scenarios encompass a wide spectrum of risky behaviors, including speeding, wrong-lane driving, aggressive merging, and head-on overtaking,providing a comprehensive and challenging benchmark for safety validation under long-tail conditions.

\subsection{Qualitative Analysis of Generated Risk Scenarios}

\begin{figure}[!t]
    \centering
    \vspace{-3mm} 
    \includegraphics[width=0.92\linewidth]{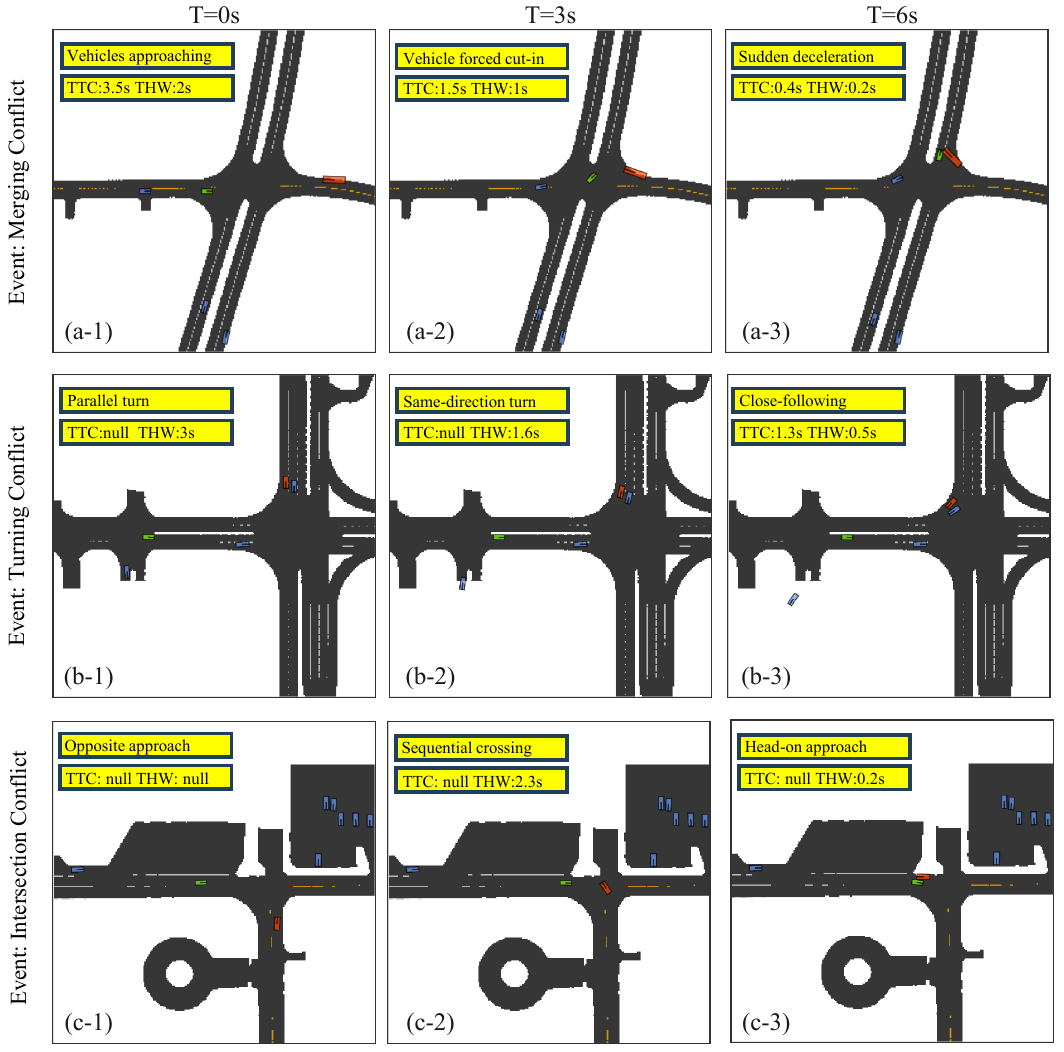}
    \vspace{-2mm}
    \caption{Evolution of vehicle interactions in long-tail traffic scenarios. 
    Rows correspond to conflict types: (a) merging, (b) turning, and (c) intersection; columns show temporal evolution at $T=0$, $3$, and $6$ s. 
    TTC and THW illustrate the progression from safe to near-collision states, highlighting the framework’s ability to generate dynamic, risk-sensitive interactions.}
    \label{fig:result}
    \vspace{-3mm}
\end{figure}

Figure~\ref{fig:result} presents representative long-tail scenarios generated by the proposed LLM-guided framework, showing how vehicle interactions evolve across different conflict types and risk levels. The visualization includes three categories—merging, turning, and intersection conflicts—each evolving over time at $T=0$ s, $3$ s, and $6$ s. Changes in TTC and THW clearly illustrate how normal interactions gradually escalate into near-collision states.
(a) Merging conflict: Initially ($T=0$ s), two vehicles maintain moderate safety margins (TTC $\approx$ 3.5 s, THW $\approx$ 2 s). As one performs an unexpected merge, TTC and THW decrease to 1.5 s and 1 s, and eventually to 0.4 s and 0.2 s during forced cut-in and deceleration, marking a clear transition from safe following to high-risk interaction.  
(b) Turning conflict: Two vehicles execute parallel left turns at an intersection. Although TTC is undefined due to lateral motion, THW decreases from about 3 s to 1.6 s, reflecting intensified proximity caused by limited visibility and right-of-way ambiguity.  
(c) Intersection conflict: Multiple vehicles converge through opposite, sequential, and head-on paths. As they approach the intersection center, TTC and THW drop sharply (e.g., THW $\approx$ 0.2 s), leaving minimal reaction time and forming near-collision conditions.
Overall, these results demonstrate that the proposed framework can reproduce realistic temporal risk escalation while maintaining spatial and kinematic consistency. The ability to synthesize such evolving, high-risk scenarios provides strong qualitative evidence of its effectiveness for safety-critical evaluation of ADS.

\begin{figure}[!t]
    \centering
    \includegraphics[width=1.0\linewidth]{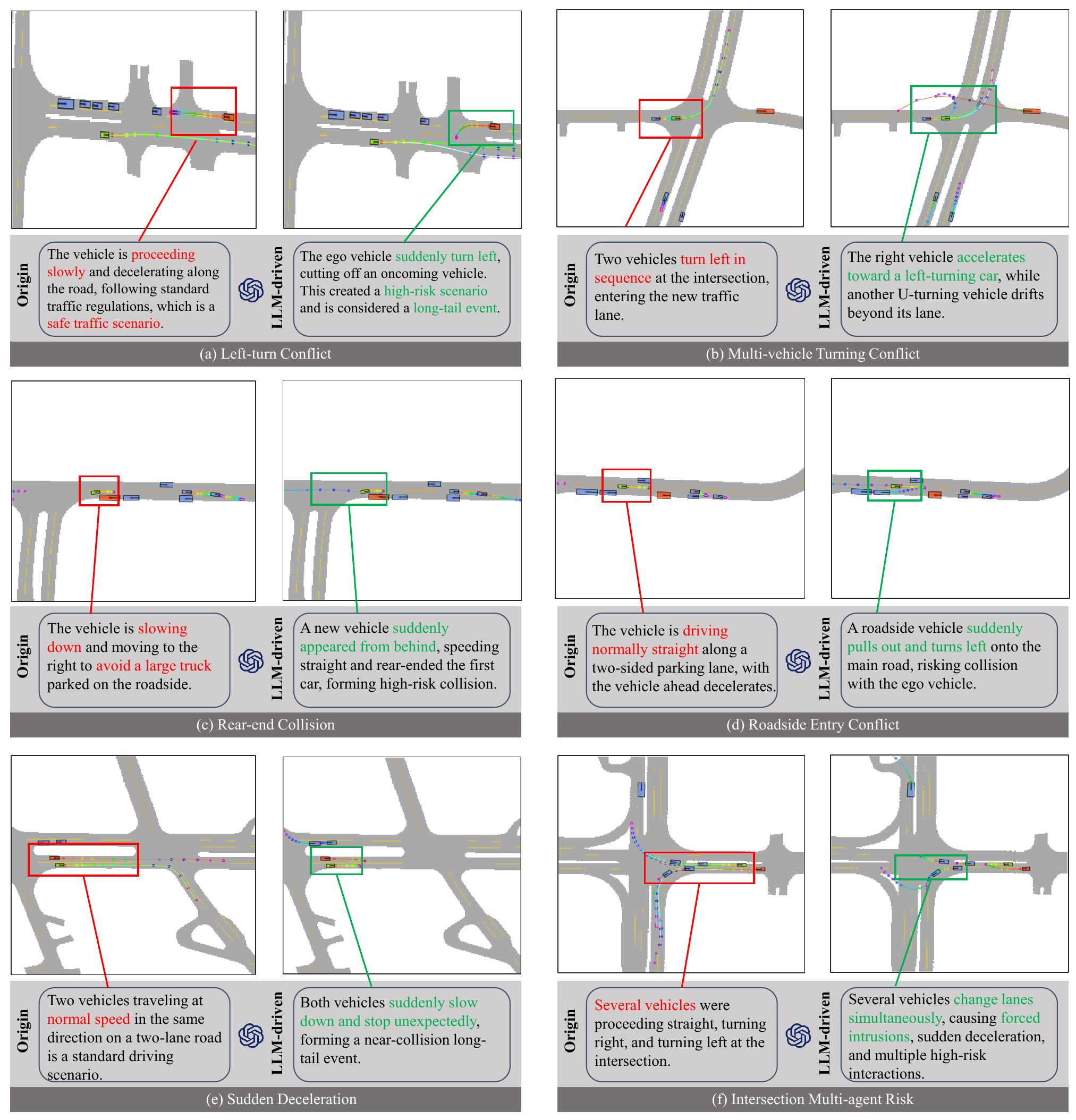}
    \caption{Comparison between original and LLM-driven scenarios. Each pair shows how the framework transforms safe trajectories (left) into high-risk, long-tail events (right): (a) left-turn conflict; (b) multi-vehicle turning conflict; (c) rear-end collision; (d) roadside entry conflict; (e) sudden deceleration; (f) intersection multi-agent risk. The LLM-guided process interprets scenario semantics and adaptively optimizes risk objectives to reproduce realistic and diverse safety-critical interactions.}
    \label{fig:compare}
\end{figure}

Figure~\ref{fig:compare} shows representative comparisons between original safe trajectories and their LLM-generated long-tail counterparts. Each pair (left: origin, right: LLM-driven) illustrates how the framework transforms normal driving behaviors into safety-critical interactions by interpreting scenario semantics and adaptively optimizing risk-sensitive objectives.
Scenarios (a–c) represent the transition from low- to high-risk conditions. In (a), the ego vehicle, initially slowing along the lane, suddenly turns left and cuts off an oncoming vehicle, forming a left-turn conflict. In (b), two vehicles turning left sequentially are disrupted by a right-side vehicle accelerating and a U-turning vehicle drifting beyond its lane, creating a multi-vehicle turning conflict. In (c), a vehicle slowing to avoid a parked truck is rear-ended by a fast-approaching vehicle, forming a high-risk collision.  
Scenarios (d–f) show more complex multi-agent interactions. In (d), a parked vehicle suddenly starts and turns left into traffic, generating a roadside entry conflict. In (e), two vehicles traveling in the same direction simultaneously decelerate and stop, forming a near-collision event. In (f), several vehicles at an intersection change lanes and turning paths simultaneously, causing forced intrusions, abrupt deceleration, and overlapping risks.  
These results demonstrate that the LLM-guided framework can evolve safe trajectories into diverse, high-risk situations while maintaining behavioral realism and physical feasibility, offering a broad range of scenarios for evaluating ADS safety under rare but critical interactions.

\begin{figure}[!t]
    \centering
    \includegraphics[width=0.95\linewidth]{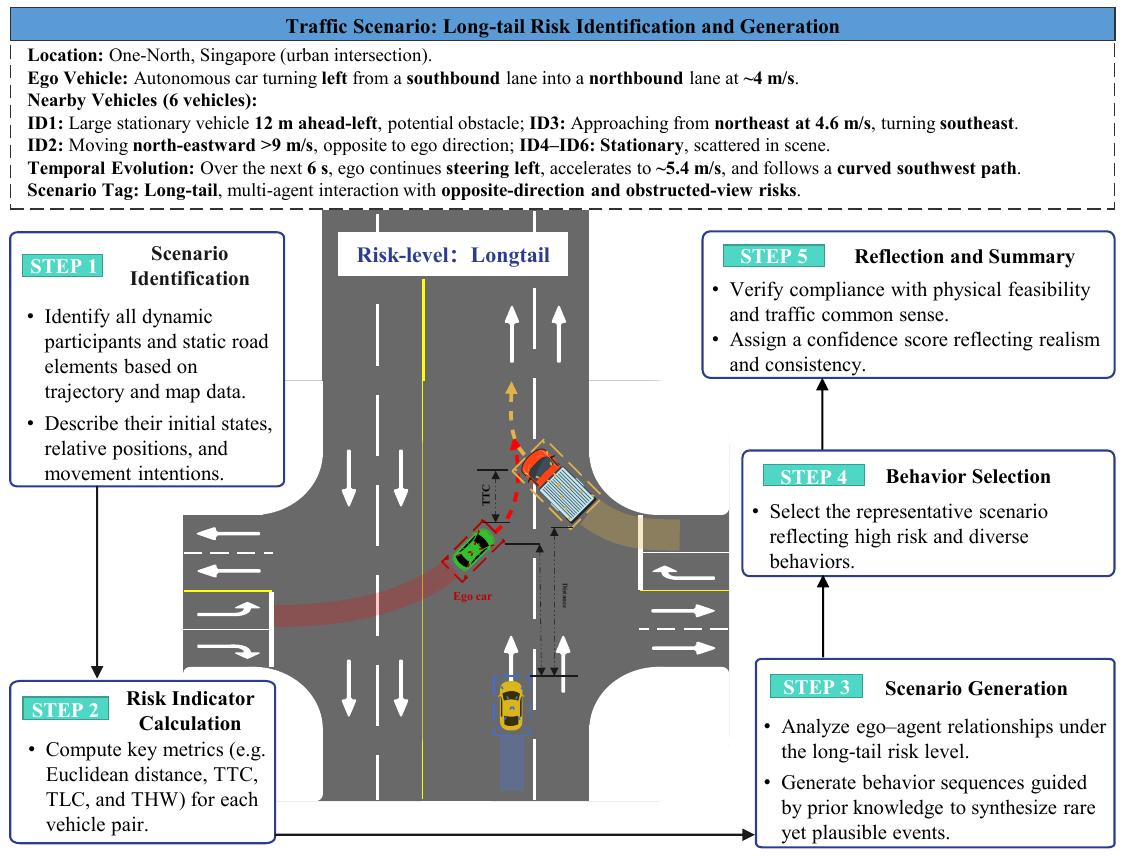}
    \caption{Case study of long-tail risk identification and generation at the One-North urban intersection in Singapore. The ego vehicle performs a left turn under multi-agent interactions involving both stationary and moving vehicles. The bottom pipeline shows five steps: (1) scenario identification, (2) risk indicator computation, (3) behavior selection, (4) LLM-guided scenario generation, and (5) validation. The process combines quantitative metrics (TTC, THW, TLC, distance) with reasoning to produce realistic long-tail traffic scenarios.}
    \label{fig:case_study}
\end{figure}

Figure~\ref{fig:case_study} presents a case study demonstrating the long-tail risk identification and scenario generation process of the proposed LLM-guided framework. The scenario is located at the One-North urban intersection in Singapore, where the ego vehicle executes a left turn while interacting with six surrounding vehicles under partial occlusion. Initially, the ego travels southbound at around 4 m/s, facing a stationary obstacle on the left and an oncoming vehicle from the northeast at 4.6 m/s. As the interaction evolves, the ego accelerates to 5.4 m/s, producing complex spatiotemporal coupling among turning, opposing, and stationary agents.
The lower part of the figure illustrates the five-step reasoning pipeline: (1) identifying all dynamic and static participants; (2) computing key indicators such as TTC, THW, TLC, and distance; (3) analyzing ego–agent relations and selecting long-tail cases; (4) generating behaviors under LLM guidance to synthesize plausible high-risk interactions; and (5) validating physical and semantic consistency. 
This case study highlights how the framework integrates quantitative metrics with language-based reasoning to reproduce rare yet realistic long-tail interactions, bridging data-driven understanding with interpretable, controllable scenario synthesis.

\begin{figure}[!t]
    \centering
    \includegraphics[width=1.0\linewidth]{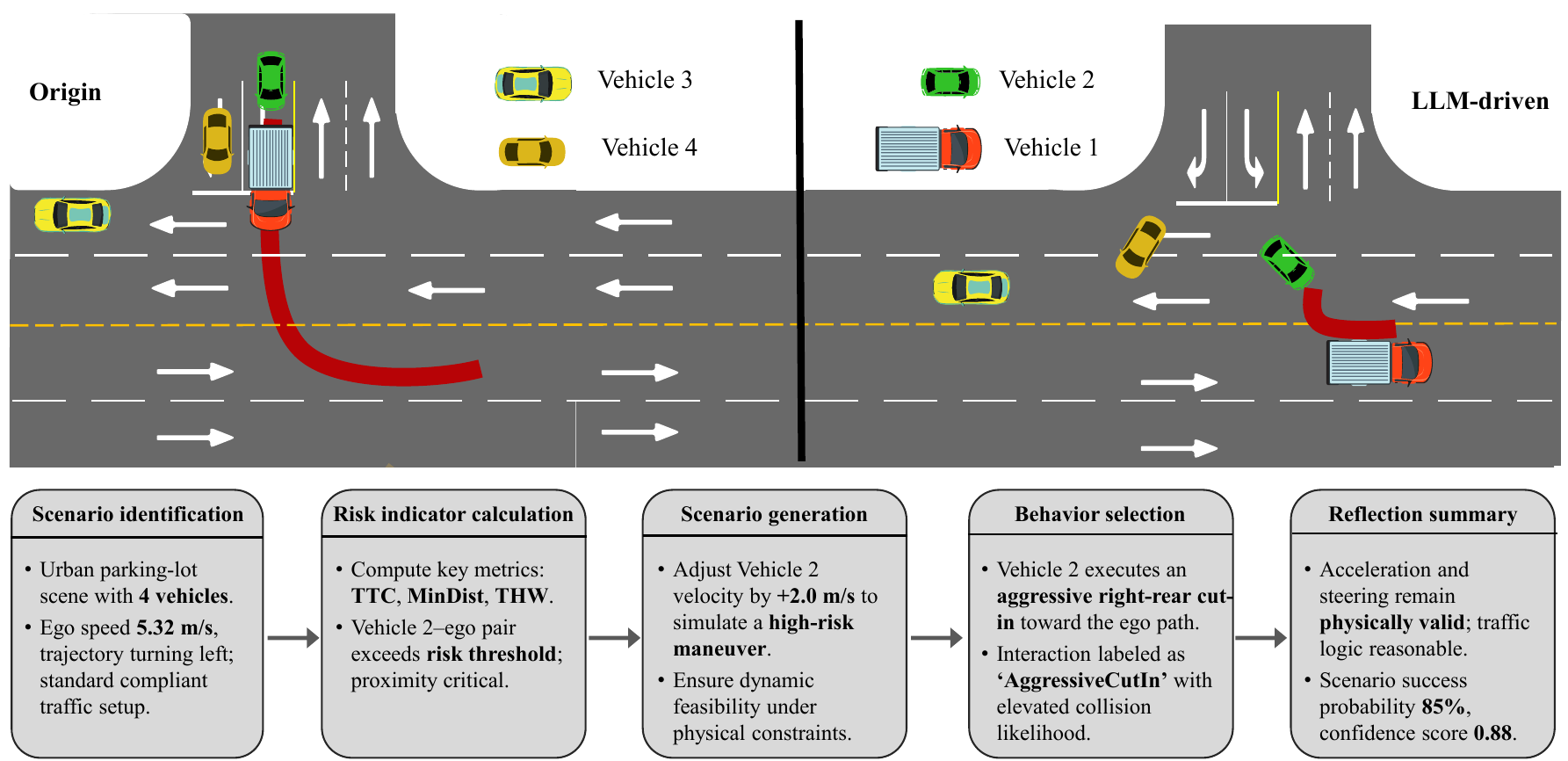}
    \caption{LLM-guided long-tail scenario generation in an urban parking-lot environment. The original scene features four vehicles in compliant low-speed motion. The framework identifies a high-risk ego–Vehicle 2 pair via quantitative indicators (TTC, minimum distance, THW) and adjusts Vehicle 2’s velocity by +2.0 m/s to simulate an aggressive right-rear cut-in. The resulting interaction remains physically feasible and semantically valid, achieving 85\% scenario success probability and 0.88 confidence score, exemplifying controllable synthesis of safety-critical behaviors.}
    \label{fig:parking_case}
\end{figure}

Figure~\ref{fig:parking_case} illustrates an LLM-guided generation example in an urban parking-lot environment, demonstrating how a normal, low-risk situation can evolve into a high-risk aggressive cut-in scenario through semantic reasoning and risk-sensitive optimization. The original scene involves four vehicles moving at low speed in compliance with standard traffic regulations. The ego vehicle travels at approximately 5.3 m/s and begins a left-turn trajectory, while surrounding vehicles maintain moderate spacing.

In Step 1, the system identifies all dynamic participants and static elements, constructing a structured representation of the scene. Step 2 computes key risk indicators including TTC, minimum distance, and THW. The ego–Vehicle 2 pair exceeds the defined risk threshold, indicating a potential hazard. In Step 3, the LLM adjusts Vehicle 2’s velocity by +2.0 m/s to simulate a more aggressive maneuver while ensuring physical feasibility and adherence to traffic logic. As a result, Vehicle 2 performs a right-rear cut-in toward the ego vehicle’s path, creating a critical proximity condition.
Finally, the generated behavior is labeled as an “AggressiveCutIn” event, with a scenario success probability of 85\% and a confidence score of 0.88. This case confirms that the LLM-guided reasoning not only enhances scenario realism but also enables controllable synthesis of safety-critical interactions that preserve kinematic validity and contextual consistency. The resulting scene offers an effective benchmark for evaluating the responsiveness and safety margin of ADS under constrained, near-collision conditions.

\FloatBarrier

\section{Conclusion}

This paper presented \textit{Learning from Risk}, a high-fidelity framework for generating safety-critical traffic scenarios by integrating a CVAE with a LLM. Addressing the scarcity of rare long-tail events and complex multi-agent interactions in naturalistic datasets, the framework unifies data-driven structure learning with knowledge-guided reasoning to achieve physically consistent, behaviorally realistic, and risk-sensitive scenario generation. The CVAE-GNN module learns latent traffic dynamics from the highD and nuScenes datasets, enabling the synthesis of diverse base scenarios across both highway and urban environments. Building upon these, the LLM parses scenario descriptions, analyzes interaction patterns, and adaptively adjusts loss functions to produce scenarios that systematically span multiple risk levels.
Experiments in CARLA and SMARTS demonstrate that the proposed framework achieves low reconstruction error and high distributional fidelity for base trajectories, while significantly improving long-tail event coverage and generating diverse adversarial interactions such as aggressive cut-ins, unsafe merges, and near-collision encounters. Compared with baseline methods, the framework consistently produces more realistic and challenging situations that expose potential vulnerabilities of ADS, providing a principled tool for risk-sensitive safety validation.
Future work will extend the framework toward real-time closed-loop testing and incorporate environmental factors such as weather and road conditions to model long-tail events under more diverse settings. In addition, mitigating potential physical or semantic hallucinations in the LLM-driven weighting process will be explored to ensure trajectory realism and enhance the framework’s reliability across heterogeneous traffic domains.

\section*{Author Contributions}

Yuhang Wang: Software, Visualization, Validation, Writing -- review \& editing.
Heye Huang: Conceptualization, Methodology, Formal analysis, Writing -- original draft, Visualization.
Zhenhua Xu: Formal analysis, Writing -- review \& editing.
Kailai Sun: Investigation, Visualization, Validation.
Baoshen Guo: Data curation, Investigation, Writing -- review \& editing.
Jinhua Zhao: Supervision, Conceptualization, Writing -- review \& editing.

\section*{Declaration of competing interest}

The authors declare that they have no known competing financial interests or personal relationships that could have appeared to influence the work reported in this paper.

\section*{Data availability}

Data will be made available on request.


\bibliography{ref}

\end{document}